\documentclass[letterpaper]{article} %
\usepackage[preprint]{aaai2027}
\usepackage[hyphens]{url}  %
\usepackage{graphicx} %
\usepackage{natbib}  %
\usepackage{caption} %
\usepackage{algorithm}
\usepackage{algorithmic}

\usepackage{newfloat}
\usepackage{listings}
\DeclareCaptionStyle{ruled}{labelfont=normalfont,labelsep=colon,strut=off} %
\floatstyle{ruled}
\newfloat{listing}{tb}{lst}{}
\floatname{listing}{Listing}

\usepackage{booktabs}

\usepackage{amsmath}
\usepackage{amssymb}
\usepackage{multirow}
\usepackage{colortbl}
\usepackage{pifont}

\newcommand{\cmark}{\ding{51}}
\newcommand{\xmark}{\ding{55}}

\title{Rethinking Fixed Temporal Grids: Frequency-Disentangled Motion Generation}
\author{
    Yunjiao Zhou,
    Junlang Qian,
    Gen Li,
    Xinying Guo,
    Lihua Xie,
    Jianfei Yang
}
\affiliations{
    Nanyang Technological University, Singapore
}

\begin{document}

\maketitle

\begin{abstract}
Most human motion generation methods encode motion as tokens on a uniform temporal grid, where every token spans the same fixed time window.
Human motion, however, is temporally heterogeneous: slowly evolving global trajectories coexist with rapid transient events such as foot contacts and joint impulses.
Forcing such multi-scale dynamics onto tokens of identical temporal resolution entangles motion frequencies, leaving slow regions redundant while smoothing out the rapid details that distinguish realistic motion.
We propose \textbf{FreqMo}, a scale-adaptive motion representation that decomposes motion into wavelet frequency bands, separating dynamics across temporal scales while preserving temporal localization and exact reconstruction.
Unified Frequency Residual Quantization (UFRQ) then encodes all bands within a single shared codebook, compressing the token sequence threefold and enabling stable single-stage generation.
Experiments show FreqMo attains SOTA fidelity with substantially improved high-frequency preservation, and the same decomposition transfers to continuous diffusion backbones.

\end{abstract}

\section{Introduction}
\label{sec:intro}

\begin{figure}[!tbp]
    \centering
    \includegraphics[width=1.0\linewidth]{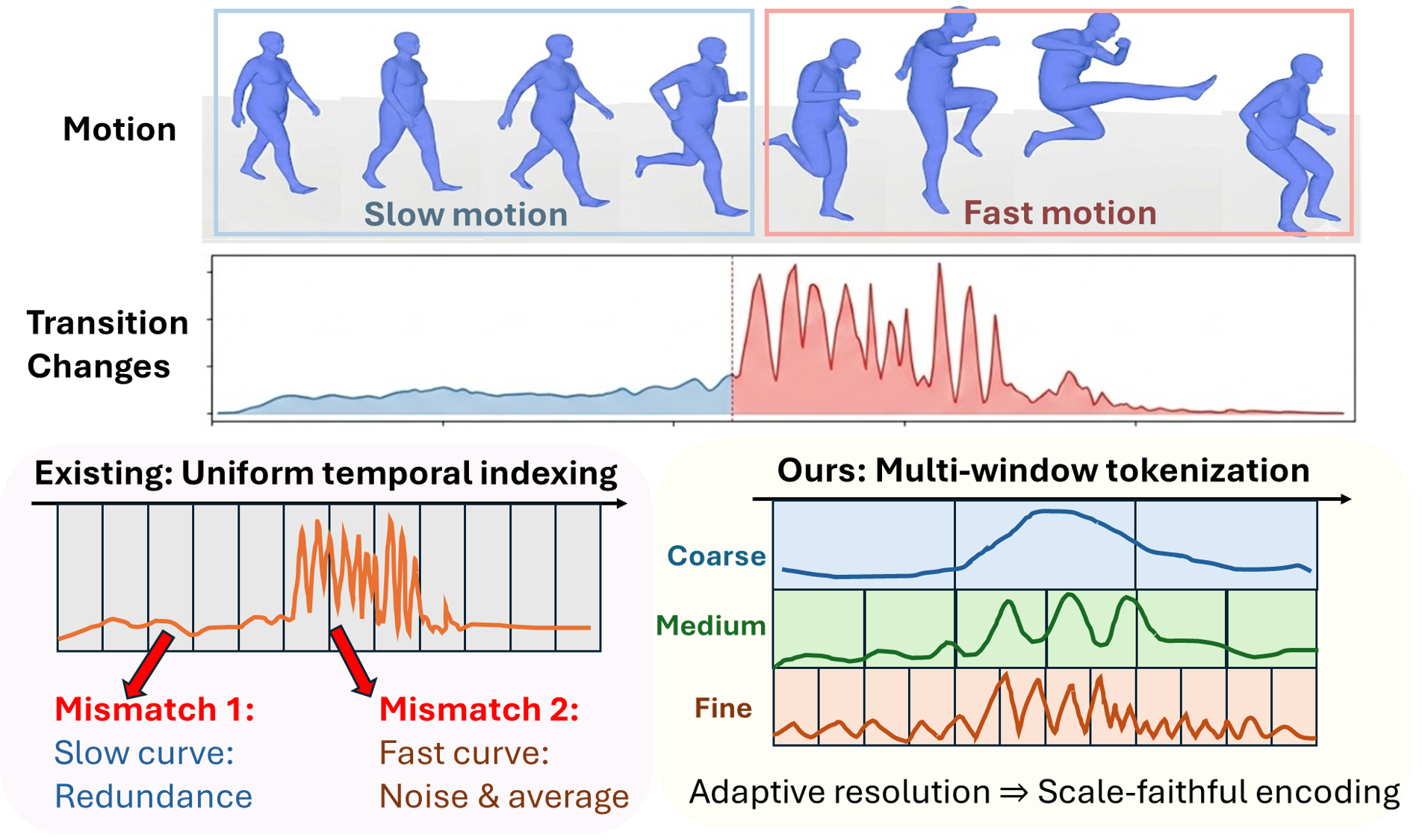}
    \vspace{-5mm}
    \caption{Left: Uniform grids induce temporal over-smoothing or redundancy. Right: FreqMo resolves it by adapting window sizes to intrinsic frequency scales.} 
    \label{fig:top}
    \vspace{-7mm}
\end{figure}

Realistic human motion generation~\cite{khani2025motionsurvey,xiao2025motionstreamer} underpins virtual reality~\cite{burdea2003virtual}, embodied intelligence~\cite{duan2022survey}, and digital human synthesis~\cite{varol2017learning}. Following language and vision~\cite{achiam2023gpt, peebles2023scalable}, prevailing methods encode motion on a fixed temporal grid where every token spans identical temporal support~\cite{cho2025discord, dai2024motionlcm, lu2025scamo}. This mirrors words in text or patches in images, where semantics reside within individual units and uniform discretization is well justified. Human motion, by contrast, is a continuous, non-stationary process whose semantics arise not from isolated poses but from temporally extended interactions: the evolution of a stride, the impulse of a landing, the deceleration into rest. Despite this distinction, few works question whether a fixed grid inherited from other modalities suits the dynamic nature of human motion (Figure~\ref{fig:top}).

This oversight has concrete representational consequences. The inter-frame dynamics that define motion semantics are temporally heterogeneous, spanning orders of magnitude from slowly evolving global trajectories to near-instantaneous events such as foot contacts and joint impulses. 
A common practice augments each token with temporal derivatives (velocity, acceleration)~\cite{guo2022generating}, yet this captures higher-order statistics of the same entangled mixture rather than disentangling its frequency components. 
Since dynamics at distinct temporal scales correspond to distinct spectral frequencies, projecting them onto identically supported tokens forces these components to mix inseparably, leading to \textbf{frequency entanglement} (Figure~\ref{fig:intro})~\cite{yuan2024mogents, fu2024mogo}. Tokens spanning slow dynamics carry redundant information, while those spanning rapid transients smooth out the details they should preserve. This distortion originates at the representation level: a structural consequence of single-resolution indexing that no per-token enrichment can resolve, not a capacity deficit.

Disentangling these frequencies is not as simple as transforming motion into the frequency domain. First, decomposing motion by frequency alone, as a global spectral transform does~\cite{fourier1822theorie}, records which frequencies occur but not when, untenable for motion, whose semantics are defined by precisely when events happen. A natural remedy reintroduces time through local windows~\cite{allen1977unified}, but at a fixed scale: one window resolves all frequencies alike, too coarse for fast transients and too fine for slow trends. What motion needs is thus resolution that adapts to frequency, together with temporal localization. On this basis, we adopt the Wavelet Transform~\cite{mallat1999wavelet}, whose variable-resolution filtering gives broad context to low-frequency trajectories and fine precision to high-frequency transients while preserving temporal resolution. This makes frequency disentanglement a property of the representation rather than the architecture.

However, directly quantizing multi-scale wavelet features inevitably introduces quantization error. A residual scheme mitigates this, encoding each band against the reconstruction of the earlier ones. The order of this refinement, however, is decisive: the wavelet hierarchy runs from high-frequency detail to coarse global structure, so refining in that order would fix transient details before the low-frequency structure that anchors the motion. We instead encode coarse to fine, locking the global structure first so that residual error is confined to the high-frequency detail where it least matters. 

Building upon this, we present \textbf{FreqMo}, a scale-adaptive motion generation framework. FreqMo decomposes motion latents into hierarchical frequency bands via multi-level wavelet transform, aligning the representation with the intrinsic temporal scale of each dynamic component. To instantiate this in discrete generative models, we propose Unified Frequency Residual Quantization (UFRQ), which encodes all frequency bands into a single shared codebook through coarse-to-fine residual refinement, replacing fragmented multi-codebook pipelines and cutting the token sequence to one third of comparable designs. A scale-aware masked transformer~\cite{chang2022maskgit} then generates all bands simultaneously for stable single-stage generation. The decomposition is moreover agnostic to the generator: transplanting it into a continuous diffusion pipeline improves that model as well, showing the gain lies in the representation rather than any particular quantization scheme.

The main contributions are summarized as follows:

\begin{itemize}
\item We identify single-scale temporal rigidity as a structural limitation in existing motion representations and characterize the resulting frequency entanglement.
\item We propose Unified Frequency Residual Quantization, which encodes wavelet-disentangled bands into a single shared codebook while preserving temporal localization and exact reconstruction, replacing multi-stage residual pipelines with stable single-stage generation.
\item We show that the same representation improves generation quality under both discrete and continuous paradigms, confirming frequency disentanglement acts at the representation level and transfers across architectures.
\end{itemize}
\section{Related Work}
\label{sec:relat}

\subsection{Motion Latent Representation}
\label{sec:rw_representation}
Learning an effective motion representation is a foundational challenge~\cite{liu2022investigating, siarohin2021motion, zhu2023motionbert}. One major paradigm utilizes continuous latent spaces, typically via Variational Autoencoders (VAEs)~\cite{kingma2013auto, chen2023executing, guo2022generating}, with recent methods operating directly on deterministic or causal latents~\cite{meng2025mardm, xiao2025motionstreamer, yu2026cmdm}. However, the high precision of the continuous space can be detrimental, forcing the model to expend capacity encoding irrelevant, high-frequency noise and minor signal variations. This hinders generalization and complicates the subsequent generative task. An alternative paradigm learns a discrete representation using Vector Quantization (VQ-VAE)~\cite{oord2017neural}. By mapping poses to a finite codebook, VQ acts as a powerful regularizer, forcing the model to discard noise and learn a robust, semantic pose vocabulary~\cite{zhang2023generating, jiang2023motiongpt,lu2025scamo}. Given this robustness, our work builds upon the discrete quantization paradigm.

\subsection{Motion Generative Models}
\label{sec:rw_generation}
Given a latent representation, a generative model is required. Diffusion Models~\cite{ho2020denoising, meng2025mardm, weng2026realign} achieve strong quality by iteratively denoising a sequence, but suffer from computational overhead and slow inference. Autoregressive models~\cite{vaswani2017attention, lu2025scamo, yu2026cmdm, bian2025motioncraft} generate tokens sequentially; while strong at capturing temporal dependencies, they are also slow and can accumulate errors. Multimodal and whole-body variants further extend this paradigm~\cite{xu2026omnimotionx, zhang2026motionvla}. To balance these trade-offs, non-autoregressive masked transformers~\cite{chang2022maskgit, pinyoanuntapong2024bamm, zeng2025lightt2m}, inspired by BERT~\cite{devlin2019bert}, have emerged as a powerful alternative. By predicting masked tokens in parallel, these models enable fast inference while maintaining high quality via iterative refinement. Therefore, we build our generator upon the bidirectional masked transformer framework.
\section{Method}
\label{sec:method}
\begin{figure*}[htbp]
    \centering
    \includegraphics[width=\linewidth]{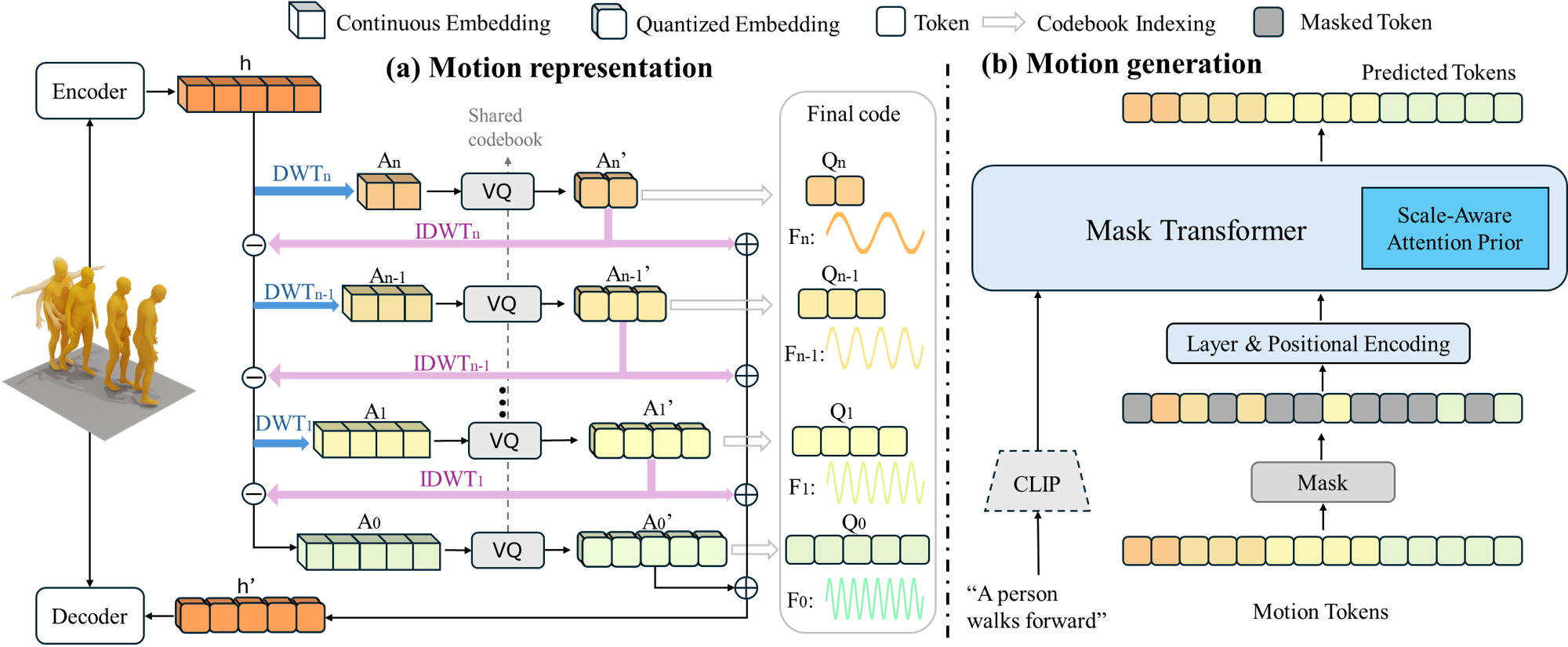}
    \vspace{-5mm}
    \caption{Model structure for FreqMo during training.}
    \label{fig:train}
    \vspace{-2mm}
\end{figure*}

Given a text description $\mathbf{c}$, our goal is to generate a motion sequence $\mathbf{X} = \{\mathbf{x}_t\}_{t=1}^T$, where each $\mathbf{x}_t \in \mathbb{R}^J$ denotes the pose at frame $t$.
Conventional representations discretize motion on a uniform temporal grid, forcing dynamics at different temporal scales to share identical support and entangling them before generation begins.
To resolve this, we introduce \textbf{FreqMo}, separating motion across temporal scales through two components: a frequency-disentangled representation that decomposes latent motion into multi-scale bands via wavelet transform and encodes them through UFRQ; and a text-conditioned masked transformer that generates all frequency tokens jointly while modeling their cross-scale dependencies.
The wavelet decomposition is the core of FreqMo, and because it is independent of the generator, it applies to both continuous and discrete latent spaces.

\subsection{Frequency-Disentangled Motion Representation}
\label{sec:representation}

A motion representation compresses raw motion into a compact, structured form that facilitates generation. Given a motion sequence $\mathbf{X} \in \mathbb{R}^{T \times J}$, a motion encoder $E_M(\cdot)$ projects it into continuous latent embeddings $\mathbf{h} \in \mathbb{R}^{T' \times d}$, where $T'$ is the downsampled temporal length and $d$ the embedding dimension. However, since $E_M$ operates on a uniform temporal grid, $\mathbf{h}$ conflates slowly evolving global trajectories and rapid local transients within the same fixed-resolution space. To resolve this, we decompose $\mathbf{h}$ into a hierarchy of frequency-disentangled bands prior to any sequence modeling or quantization, ensuring that multi-scale dynamics are structurally separated at the representation level.

\subsubsection{Wavelet-Based Frequency Decomposition.}
\label{sec:decomposition}

The decomposition operates through the discrete wavelet transform (DWT), applying a pair of complementary filters to separate frequency components. At decomposition level $l$, given an input $\mathbf{X}_l \in \mathbb{R}^{T_l \times d}$ with temporal length $T_l$, DWT produces:
\begin{equation}
\mathbf{A}_l, \mathbf{D}_l = \text{DWT}(\mathbf{X}_l),
\end{equation}
where $\mathbf{A}_l \in \mathbb{R}^{T_{l+1} \times d}$ is the approximation capturing smooth, low-frequency structure, and $\mathbf{D}_l \in \mathbb{R}^{T_{l+1} \times d}$ the detail capturing rapid, high-frequency variations. Both are downsampled from $T_l$ to $T_{l+1} < T_l$.

Starting from the initial latent embeddings $\mathbf{X}_1 = \mathbf{h} \in \mathbb{R}^{T' \times d}$ obtained from the motion encoder, DWT produces the first-level decomposition. However, the approximation $\mathbf{A}_1$ still contains frequency variations that oscillate too slowly to be captured at the current temporal resolution. By recursively applying DWT to the approximation, setting $\mathbf{X}_{l+1} = \mathbf{A}_l$, we progressively isolate slower frequency components at coarser temporal scales. After $L$ levels, this yields a hierarchy of frequency bands:
\begin{equation}
\mathbf{A}_L \in \mathbb{R}^{T_L \times d}, \quad \{\mathbf{D}_l \in \mathbb{R}^{T_{l+1} \times d}\}_{l=1}^L,
\end{equation}
where 
lower-indexed bands ($\mathbf{D}_1$) encode finer, higher-frequency dynamics at higher temporal resolution and higher-indexed bands ($\mathbf{D}_L$) encode coarser, lower-frequency changes at lower resolution. This multi-scale hierarchy naturally emerges from the recursive decomposition, with temporal resolution decreasing as $T_1 > T_2 > \cdots > T_L$.

Crucially, the wavelet transform guarantees perfect reconstruction via inverse DWT (IDWT):
\begin{equation}
\mathbf{X}_l = \text{IDWT}(\mathbf{A}_l, \mathbf{D}_l),
\end{equation}
so that $\hat{\mathbf{h}} = \text{IDWT}^L(\mathbf{A}_L, \{\mathbf{D}_l\}_{l=1}^L)$ recovers the original representation losslessly. This hierarchy is the frequency-disentangled representation: each band captures dynamics at a specific temporal scale, and exact invertibility ensures no kinematic information is discarded during decomposition.

\subsection{Unified Frequency Residual Quantization}
\label{sec:quantization}

Quantizing each band $\{\mathbf{A}_L, \mathbf{D}_l\}$ independently would fragment the latent space across scales and let quantization errors at coarse levels irreversibly corrupt finer details. 
UFRQ avoids both by encoding bands from coarse to fine within a single shared codebook, progressively capturing reconstruction residuals. Starting from the coarsest band $\mathbf{A}_L$ (Algorithm~\ref{alg:encode}), we iteratively quantize $\mathbf{A}_l$, reconstruct a partial signal via IDWT (with high-frequency components zeroed), and compute the residual against the current signal. This residual is then decomposed to the next finer approximation $\mathbf{A}_{l-1}$ and quantized in the same way. All frequency bands share a single codebook $\mathcal{C}$, as wavelet decomposition naturally aligns coefficient magnitudes across levels. During decoding (Algorithm~\ref{alg:decode}), quantized approximations are reconstructed in reverse order and accumulated, with the final residual added to recover the full signal. This coarse-to-fine encoding order ensures 
low-frequency structure is captured with high fidelity first, while quantization errors are relegated to less critical high-frequency bands through progressive residual refinement.

\begin{figure*}[htbp]
    \centering
    \includegraphics[width=0.8\linewidth]{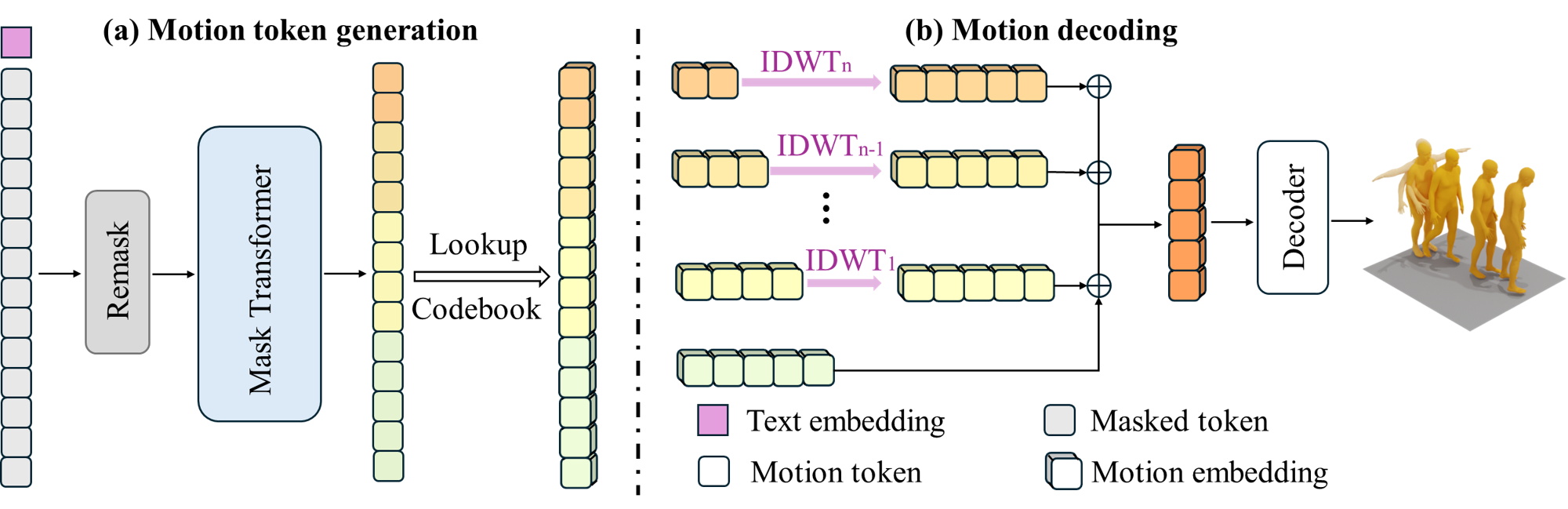}
    \vspace{-3mm}
    \caption{Model structure for FreqMo during inference.}
    \label{fig:eval}
    \vspace{-3mm}
\end{figure*}

\begin{algorithm}[t]
\small
\caption{Frequency-Residual Encoding}
\label{alg:encode}
\begin{algorithmic}[1]
\STATE \textbf{Inputs:} latent embeddings $\mathbf{h}$
\STATE \textbf{Hyperparameters:} levels $L$, codebook $\mathcal{C} \in \mathbb{R}^{K \times d}$
\STATE $\mathbf{A}_L, \{\mathbf{D}_l\}_{l=1}^L = \text{DWT}^L(\mathbf{h})$
\STATE $\mathbf{R} = \mathbf{h}$; \quad tokens $= []$
\FOR{$l = L$ \textbf{down to} $1$}
    \STATE $z_l = \text{Quantize}(\mathbf{A}_l, \mathcal{C})$ \COMMENT{Quantize frequency band}
    \STATE tokens.append$(z_l)$
    \STATE $\hat{\mathbf{A}}_l = \text{Lookup}(z_l, \mathcal{C})$ \COMMENT{Retrieve from codebook}
    \STATE $\mathbf{S}' = \text{IDWT}^l(\hat{\mathbf{A}}_l, \mathbf{0})$ \COMMENT{Partial reconstruction; $D_l$ is captured in residual}
    \STATE $\mathbf{R} = \mathbf{R} - \mathbf{S}'$ \COMMENT{Compute residual}
    \IF{$l > 1$}
        \STATE $\mathbf{A}_{l-1}, \{\mathbf{D}_j\}_{j=1}^{l-1} = \text{DWT}^{l-1}(\mathbf{R})$
    \ELSE
        \STATE $z_0 = \text{Quantize}(\mathbf{R}, \mathcal{C})$
        \STATE tokens.append$(z_0)$
    \ENDIF
\ENDFOR
\STATE \textbf{Return:} tokens $= [z_L, z_{L-1}, \ldots, z_1, z_0]$

\end{algorithmic}
\end{algorithm}

\begin{algorithm}[t]

\caption{Frequency-Residual Decoding}
\label{alg:decode}
\begin{algorithmic}[1]
\small
\STATE \textbf{Inputs:} tokens $[z_L, z_{L-1}, \ldots, z_1, z_0]$
\STATE \textbf{Hyperparameters:} levels $L$, codebook $\mathcal{C}$
\STATE $\hat{\mathbf{A}}_L = \text{Lookup}(z_L, \mathcal{C})$
\STATE $\mathbf{S}' = \text{IDWT}^L(\hat{\mathbf{A}}_L, \mathbf{0})$
\FOR{$l = L-1$ \textbf{down to} $1$}
    \STATE $\hat{\mathbf{A}}_l = \text{Lookup}(z_l, \mathcal{C})$
    \STATE $\mathbf{S}'' = \text{IDWT}^l(\hat{\mathbf{A}}_l, \mathbf{0})$
    \STATE $\mathbf{S}' = \mathbf{S}' + \mathbf{S}''$ \COMMENT{Accumulate reconstruction}
\ENDFOR
\STATE $\hat{\mathbf{R}} = \text{Lookup}(z_0, \mathcal{C})$
\STATE $\hat{\mathbf{h}} = \mathbf{S}' + \hat{\mathbf{R}}$
\STATE \textbf{Return:} $\hat{\mathbf{h}}$
\end{algorithmic}
\end{algorithm}

\textbf{Training Objective.}
The tokenizer is trained end-to-end to minimize reconstruction error while ensuring stable codebook utilization. The reconstructed latent $\hat{\mathbf{h}}$ is decoded by $D_M$ into motion $\hat{\mathbf{X}}$, and the full system is optimized via:
$$
\mathcal{L}_{\text{rep}} = \|\mathbf{X} - \hat{\mathbf{X}}\|_1 + \beta \sum_{i=0}^{L} \|\mathbf{v}_i - \text{sg}[\hat{\mathbf{v}}_i]\|_F^2,
$$
where $\mathbf{v}_i$ denotes pre-quantization features ($\mathbf{A}_i$ for $i \ge 1$, $\mathbf{R}$ for $i=0$), $\hat{\mathbf{v}}_i$ the corresponding codebook embedding, $\beta$ a balancing weight, and $\text{sg}[\cdot]$ the stop-gradient operator.

\subsection{Text-to-Motion Generation Model}
\label{sec:generation}

Given the pretrained representation, the task is to generate the token sequence $\mathbf{z} = [z_L, \ldots, z_1, z_0]$ conditioned on text $\mathbf{c}$ (Figure~\ref{fig:train}b). Unlike autoregressive or multi-stage methods, we use a bidirectional masked transformer~\cite{devlin2019bert} that produces all frequency bands simultaneously, reasoning globally across scales without cascaded pipelines. Text is encoded via CLIP~\cite{radford2021learning}, and a fraction $\rho$ of tokens in $\mathbf{z}$ are replaced with \texttt{[MASK]} to obtain $\tilde{\mathbf{z}}$. Each token carries a sinusoidal encoding of its normalized timestamp $\tau_i = (k{+}0.5)/N_l$ ($k$ indexing tokens within level $l$) and a learnable scale embedding $\mathbf{s}_l$ for its frequency level, aligning tokens from different resolutions that share the same physical time. A Transformer encoder then processes $\tilde{\mathbf{z}}$ via self-attention over the full sequence and cross-attention to text, predicting the original tokens at masked positions.

\subsubsection{Scale-aware attention prior.} In the multi-scale token sequence, high-frequency bands contribute more tokens and may dominate attention, diluting focus on low-frequency structure. To counteract this, we introduce a scale-aware prior that biases attention toward coarser levels. Let $N_l$ denote the token count at frequency level $l$, and $N = \sum_{l=0}^{L} N_l$ the total sequence length. Each token $i$ is associated with a frequency level $l(i)$, with coarser levels receiving higher weights $w_l = \log(T_{\max}/T_l)$ normalized across scales. These form a fixed bias matrix $\mathbf{M}_{\text{bias}} \in \mathbb{R}^{N \times N}$ with $\mathbf{M}_{\text{bias}}[i,j] = w_{l(i)} + w_{l(j)}$, added to the attention logits:
\begin{equation}
\text{Attention}(\mathbf{Q}, \mathbf{K}) = \text{softmax}\left(\frac{\mathbf{Q}\mathbf{K}^T}{\sqrt{d_k}} + \mathbf{M}_{\text{bias}}\right)\mathbf{V}.
\end{equation}
This keeps coarse structure emphasized throughout, preventing overfitting to dense high-frequency bands.

\textbf{Training Objective.} The model minimizes cross-entropy over masked positions:
\begin{equation}
\mathcal{L}_{\text{gen}} = -\mathbb{E}_{\tilde{\mathbf{z}} \sim p_{\text{mask}}(\mathbf{z})} \left[ \sum_{i \in \mathcal{M}} \log p_\theta(z_i | \tilde{\mathbf{z}}, \mathbf{c}) \right],
\end{equation}
where $\mathcal{M}$ denotes masked positions and $p_{\text{mask}}$ the masking distribution with ratio $\rho$. The tokenizer remains frozen.

\subsection{Inference}
\label{sec:inference}

At inference time (Figure~\ref{fig:eval}), motion is generated from text in two stages. Given $\mathbf{c}$, the model first generates $\mathbf{z} = [z_L, \ldots, z_0]$ via iterative refinement: starting from all \texttt{[MASK]} tokens, it performs $M$ steps, at each step predicting confidence scores, unmasking the top $k_m$ positions, and re-predicting remaining tokens. The completed sequence is decoded into $\hat{\mathbf{h}}$ via Algorithm~\ref{alg:decode} and passed to the motion decoder $D_M$ to produce the final motion $\hat{\mathbf{X}}$.

\section{Experiment}
\begin{table*}[htbp]
\centering
\resizebox{0.8\textwidth}{!}{%
\begin{tabular}{l l c ccc cc}
\toprule
\multirow{2}{*}{\textbf{Methods}} & \multirow{2}{*}{\textbf{Type}} & \multirow{2}{*}{\textbf{FID$\downarrow$}} & \multicolumn{3}{c}{\textbf{R-Precision$\uparrow$}} & \multirow{2}{*}{\textbf{MM. Dist.$\downarrow$}} & \multirow{2}{*}{\textbf{MM.$\uparrow$}} \\
\cmidrule(lr){4-6}
 &  &  & \textbf{Top1} & \textbf{Top2} & \textbf{Top3} &  &  \\
\midrule
\rowcolor{gray!15}
\multicolumn{8}{l}{\textit{On the HumanML3D dataset}} \\
\midrule
MDM~\cite{tevet2022human} & Diff. & 0.544$^{\pm.044}$ & - & - & 0.611$^{\pm.007}$ & 5.566$^{\pm.027}$ & \textbf{2.799}$^{\pm.072}$ \\
MLD~\cite{chen2023executing} & Diff. & 0.473$^{\pm.013}$ & 0.481$^{\pm.003}$ & 0.673$^{\pm.003}$ & 0.772$^{\pm.002}$ & 3.196$^{\pm.010}$ & 2.413$^{\pm.079}$ \\
MotionDiffuse~\cite{zhang2024motiondiffuse} & Diff. & 0.630$^{\pm.001}$ & 0.491$^{\pm.001}$ & 0.681$^{\pm.001}$ & 0.782$^{\pm.001}$ & 3.113$^{\pm.001}$ & 1.553$^{\pm.042}$ \\
ReMoDiffuse~\cite{zhang2023remodiffuse} & Diff. & 0.103$^{\pm.004}$ & 0.510$^{\pm.005}$ & 0.698$^{\pm.006}$ & 0.795$^{\pm.004}$ & 2.974$^{\pm.016}$ & 1.795$^{\pm.043}$ \\

\midrule
TM2T~\cite{guo2022tm2t} & VQ & 1.501$^{\pm.017}$ & 0.424$^{\pm.003}$ & 0.618$^{\pm.003}$ & 0.729$^{\pm.002}$ & 3.467$^{\pm.011}$ & \underline{2.424}$^{\pm.093}$ \\
T2M~\cite{guo2022generating} & VQ & 1.087$^{\pm.021}$ & 0.455$^{\pm.003}$ & 0.636$^{\pm.003}$ & 0.736$^{\pm.002}$ & 3.347$^{\pm.008}$ & 2.219$^{\pm.074}$ \\
T2M-GPT~\cite{zhang2023generating} & VQ & 0.141$^{\pm.005}$ & 0.492$^{\pm.003}$ & 0.679$^{\pm.002}$ & 0.775$^{\pm.002}$ & 3.121$^{\pm.009}$ & 1.831$^{\pm.048}$ \\
ParCo~\cite{zou2024parco} & VQ & 0.109$^{\pm.005}$ & 0.515$^{\pm.003}$ & 0.709$^{\pm.003}$ & 0.801$^{\pm.002}$ & \textbf{2.927}$^{\pm.008}$ & 1.382$^{\pm.060}$ \\
MoMask~\cite{guo2024momask} & VQ & 0.045$^{\pm.002}$ & \textbf{0.521}$^{\pm.002}$ & \textbf{0.713}$^{\pm.002}$ & \textbf{0.807}$^{\pm.002}$ & 2.958$^{\pm.008}$ & 1.241$^{\pm.040}$ \\
MoGo~\cite{fu2025mogo} & VQ & 0.064$^{\pm.002}$ & \underline{0.515}$^{\pm.003}$ & \underline{0.709}$^{\pm.003}$ & 0.801$^{\pm.003}$ & 2.951$^{\pm.008}$ & 2.108$^{\pm.070}$ \\
MARDM~\cite{meng2025mardm} & - & 0.114$^{\pm.007}$ & 0.500$^{\pm.004}$ & 0.695$^{\pm.003}$ & 0.795$^{\pm.003}$ & 3.270$^{\pm.009}$ & 2.231$^{\pm.071}$ \\
\midrule
FreqMo-Diff & Diff. & 0.043$^{\pm.002}$ & 0.508$^{\pm.003}$ & 0.705$^{\pm.004}$ & 0.798$^{\pm.002}$ & \underline{2.986}$^{\pm.009}$ & 1.347$^{\pm.052}$ \\
\textbf{FreqMo} & VQ & \textbf{0.040}$^{\pm.002}$ & 0.519$^{\pm.003}$ & 0.712$^{\pm.003}$ & \underline{0.802}$^{\pm.003}$ & 3.025$^{\pm.010}$ & 1.265$^{\pm.019}$ \\
\midrule
\rowcolor{gray!15}
\multicolumn{8}{l}{\textit{On the KIT-ML dataset}} \\
\midrule
MDM~\cite{tevet2022human} & Diff. & 0.497$^{\pm.021}$ & - & - & 0.396$^{\pm.004}$ & 9.191$^{\pm.022}$ & 1.907$^{\pm.214}$ \\
MLD~\cite{chen2023executing} & Diff. & 0.404$^{\pm.027}$ & 0.390$^{\pm.008}$ & 0.608$^{\pm.008}$ & 0.734$^{\pm.007}$ & 3.204$^{\pm.027}$ & \textbf{2.192}$^{\pm.071}$ \\
MotionDiffuse~\cite{zhang2024motiondiffuse} & Diff. & 1.954$^{\pm.062}$ & 0.417$^{\pm.004}$ & 0.621$^{\pm.004}$ & 0.739$^{\pm.004}$ & 2.958$^{\pm.005}$ & 0.730$^{\pm.013}$ \\
ReMoDiffuse~\cite{zhang2023remodiffuse} & Diff. & \textbf{0.155}$^{\pm.006}$ & 0.427$^{\pm.014}$ & 0.641$^{\pm.004}$ & 0.765$^{\pm.005}$ & 2.814$^{\pm.012}$ & 1.239$^{\pm.028}$ \\

\midrule
TM2T~\cite{guo2022tm2t} & VQ & 3.599$^{\pm.153}$ & 0.280$^{\pm.005}$ & 0.463$^{\pm.006}$ & 0.587$^{\pm.005}$ & 4.591$^{\pm.026}$ & \textbf{3.292}$^{\pm.081}$ \\
T2M~\cite{guo2022generating} & VQ & 3.022$^{\pm.107}$ & 0.361$^{\pm.005}$ & 0.559$^{\pm.007}$ & 0.681$^{\pm.007}$ & 3.488$^{\pm.028}$ & 2.052$^{\pm.107}$ \\
T2M-GPT~\cite{zhang2023generating} & VQ & 0.514$^{\pm.029}$ & 0.416$^{\pm.006}$ & 0.627$^{\pm.006}$ & 0.745$^{\pm.005}$ & 3.007$^{\pm.023}$ & 1.570$^{\pm.039}$ \\
ParCo~\cite{zou2024parco} & VQ & 0.453$^{\pm.027}$ & 0.430$^{\pm.004}$ & 0.649$^{\pm.007}$ & 0.772$^{\pm.006}$ & 2.820$^{\pm.028}$ & 1.245$^{\pm.022}$ \\
MoMask~\cite{guo2024momask} & VQ & 0.204$^{\pm.011}$ & 0.433$^{\pm.007}$ & 0.656$^{\pm.005}$ & 0.781$^{\pm.005}$ & 2.779$^{\pm.022}$ & 1.131$^{\pm.043}$ \\
MoGo~\cite{fu2025mogo} & VQ & 0.313$^{\pm.016}$ & 0.420$^{\pm.007}$ & 0.634$^{\pm.007}$ & 0.754$^{\pm.007}$ & 2.957$^{\pm.029}$ & 2.063$^{\pm.066}$ \\
MARDM~\cite{meng2025mardm} & - & 0.242$^{\pm.014}$ & 0.387$^{\pm.006}$ & 0.610$^{\pm.006}$ & 0.749$^{\pm.006}$ & 3.374$^{\pm.019}$ & 1.312$^{\pm.053}$ \\
\midrule
FreqMo-Diff & Diff. & \underline{0.174}$^{\pm.014}$ & 0.429$^{\pm.009}$ & 0.647$^{\pm.006}$ & 0.766$^{\pm.004}$ & 2.824$^{\pm.013}$ & 1.341$^{\pm.028}$ \\
\textbf{FreqMo} & VQ & 0.181$^{\pm.009}$ & \underline{0.437}$^{\pm.006}$ & \underline{0.664}$^{\pm.007}$ & \textbf{0.796}$^{\pm.005}$ & \underline{2.769}$^{\pm.016}$ & 1.217$^{\pm.036}$ \\
\bottomrule
\end{tabular}%
}
\vspace{-1mm}
\caption{Performance comparison on HumanML3D and KIT-ML datasets. Bold: best result, underline: second best.}
\label{tab:main}
\end{table*}

\subsection{Setup}
\textbf{Datasets.} We evaluate on two standard benchmarks: HumanML3D~\cite{guo2022generating} (14,616 motions, 44,970 texts) and KIT-ML~\cite{plappert2016kit} (3,911 motions, 6,278 descriptions), both mirrored for augmentation and split 0.8:0.05:0.15 into train/validation/test.

\noindent\textbf{Evaluation metrics.} Following T2M~\cite{guo2022generating}, we report \textit{FID} for motion quality, \textit{R-Precision} (top-1/2/3) and \textit{Multimodal Distance} for text-motion alignment, and \textit{Multimodality} for diversity, prioritizing FID and R-Precision since diversity alone can reward degenerate solutions. To assess high-frequency preservation directly, we additionally report \textit{Acceleration Error} and \textit{Jerk Error}, the fidelity of second- and third-order motion derivatives.

\noindent\textbf{Implementation details.} The encoder and decoder use residual blocks with downsampling factor 4. Wavelet decomposition uses $L=5$ levels with db2 wavelets, and the shared codebook holds $K=512$ codes of dimension $d=512$. The tokenizer is trained for 200k iterations (batch size 512, learning rate $2\times10^{-4}$, commitment weight $\beta=0.25$); the masked transformer has 8 layers, 6 heads, and hidden dimension 384, with masking ratio $\rho=0.3$. The generator is trained for 100k iterations (batch size 64) with the tokenizer frozen, and inference uses $M=12$ refinement steps. All models are implemented in PyTorch on a single NVIDIA 4090 GPU.

\subsection{Generation Performance}
\label{sec:main_table}

\noindent\textbf{Quantitative results.} Table~\ref{tab:main} compares FreqMo against state-of-the-art methods. FreqMo achieves the best FID on HumanML3D (0.040) and a competitive FID on KIT-ML (0.174). While ReMoDiffuse attains a lower FID (0.155) through retrieval augmentation, FreqMo operates without any external retrieval mechanism, making the comparison complementary rather than direct. For consistency, we compare only methods reporting averaged results with confidence intervals; single-run methods without variance (e.g., Light-T2M~\cite{zeng2025lightt2m}) are discussed in the appendix. Notably, evaluating the wavelet representation under a continuous paradigm (FreqMo-Diff), which replaces UFRQ with a VAE and diffusion decoder, consistently outperforms its frame-wise counterpart MLD~\cite{chen2023executing}, confirming the gains stem from the frequency-disentangled representation rather than the quantization scheme.

\begin{figure*}[htbp]
    \centering
    \vspace{-8mm}
    \includegraphics[width=0.8\linewidth]{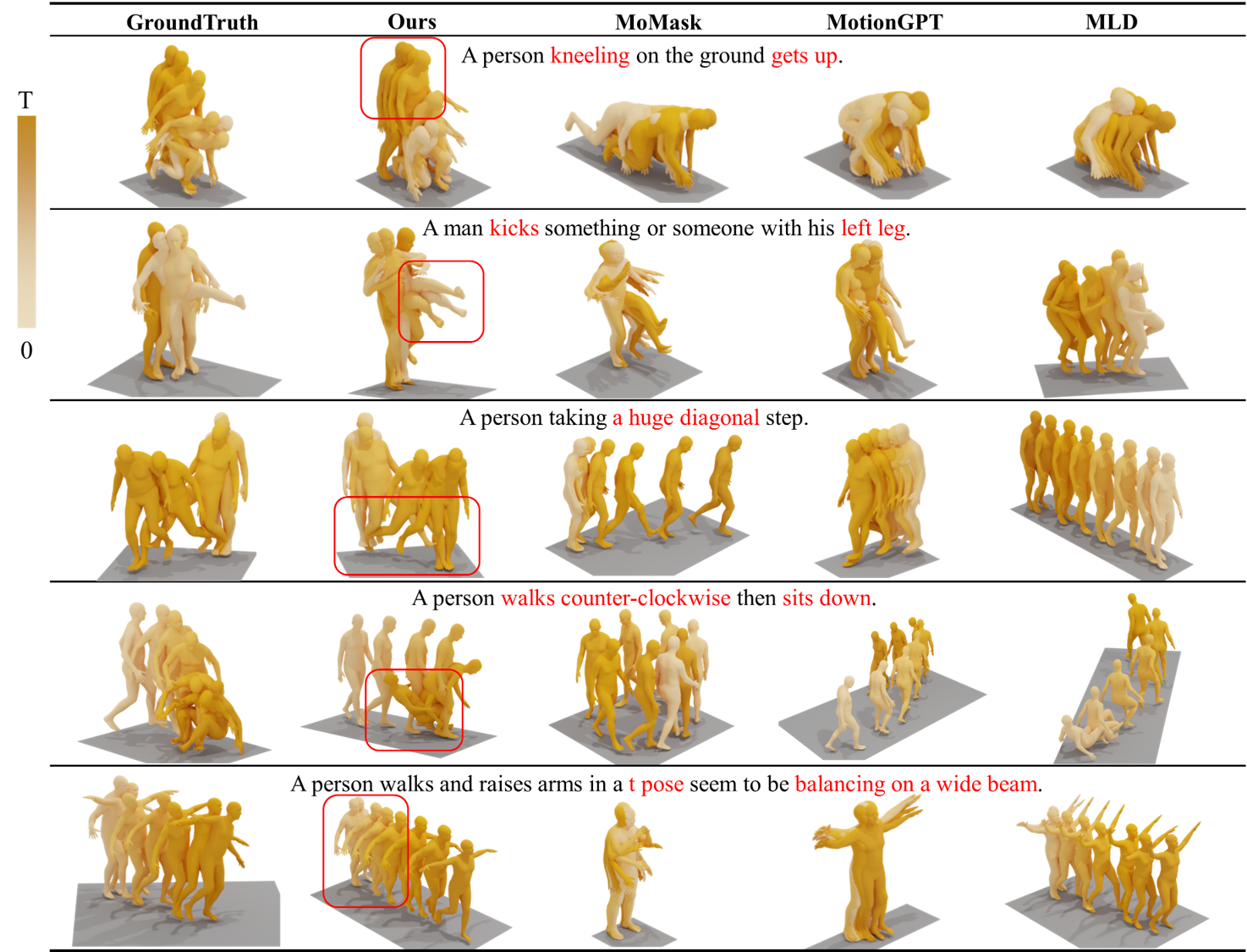}
    \vspace{-2mm}
    \caption{Visual comparisons between the different methods given the distinct text descriptions from HumanML3D testset.}
    \label{fig:visualization}
    \vspace{-5mm}
\end{figure*}

\noindent\textbf{Qualitative comparisons.} As shown in Figure~\ref{fig:visualization}, FreqMo achieves superior text-motion alignment on complex prompts, accurately rendering specific attributes (e.g., ``left leg") and compositional commands (e.g., ``walks... and... t pose") that baselines miss. See the supplementary videos for more examples.

\subsection{Representation Quality}
\label{sec:repr_quality}

\begin{table}[t]
\centering

\resizebox{\columnwidth}{!}{
\begin{tabular}{l cc c cc c}
\toprule
& \multicolumn{3}{c}{\textit{Reconstruction}} & \multicolumn{3}{c}{\textit{High-Frequency Fidelity}} \\
\cmidrule(lr){2-4} \cmidrule(lr){5-7}
\textbf{Method} & \textbf{MPJPE}$\downarrow$ & \textbf{\#Token}$\downarrow$ & \textbf{\#Codebook}$\downarrow$ & \textbf{Accel}$\downarrow$ & \textbf{Jerk}$\downarrow$ & \textbf{HF-RMSE}$\downarrow$ \\
\midrule
VQ-VAE & 58.7 & 49 & 1 & 3.42 & 854 & 0.045 \\
FSQ & 51.5 & 49 & -- & 3.88 & 911 & 0.038 \\
RQ-VAE$_2$ & 46.8 & 98 & 2 & 2.85 & 629 & 0.032 \\
RQ-VAE$_6$ & \textbf{28.7} & 294 & 6 & \underline{2.41} & \underline{586} & \underline{0.024} \\
\textbf{FreqMo} & \underline{29.5} & \textbf{106} & \textbf{1} & \textbf{2.15} & \textbf{532} & \textbf{0.012} \\
\bottomrule
\end{tabular}
}
\vspace{-2mm}
\caption{Quantization representation quality comparison.}
\label{tab:repr_quality}
\end{table}

\noindent\textbf{Reconstruction performance.} To isolate the representation from the generation model, Table~\ref{tab:repr_quality} evaluates reconstruction fidelity and high-frequency preservation across quantization methods. Compared to $\text{RQ-VAE}_6$, the strongest error-based baseline, FreqMo reduces Acceleration Error by 10.8\% (2.15 vs.\ 2.41) and Jerk Error by 9.2\% (532 vs.\ 586) while using one-third of the tokens (106 vs.\ 294) and a single shared codebook. This supports our central claim: disentangling frequencies at the representation level preserves the rapid transient dynamics that uniform encodings smooth out.

\begin{figure}[!ht]
    \centering
    \includegraphics[width=\linewidth]{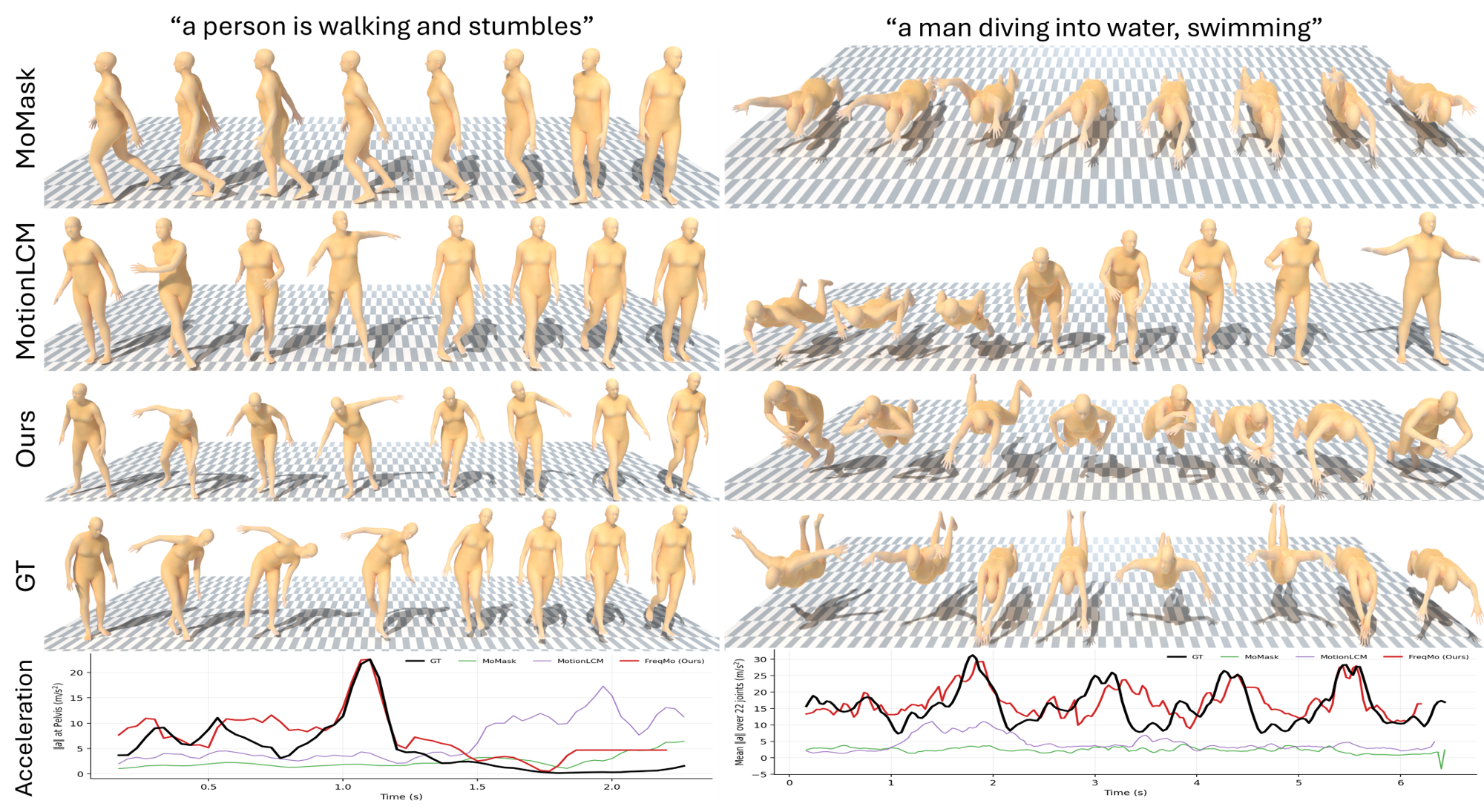}
    \vspace{-5mm}
    \caption{Side-by-side comparison on test-set prompts.}
    \vspace{-2mm}
    \label{fig:acce}
\end{figure}

\begin{figure}[t]
    \centering
    \includegraphics[width=0.82\linewidth]{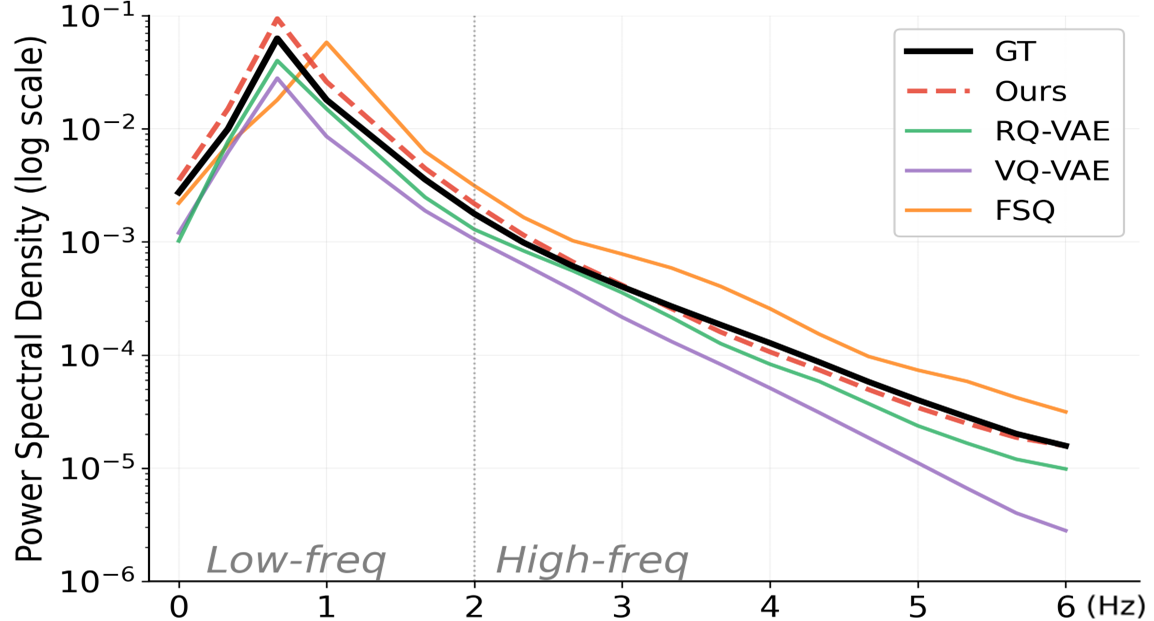}
    \vspace{-3mm}
    \caption{Power spectral density comparison.}
    \vspace{-2mm}
    \label{fig:psd}
\end{figure}

\begin{figure*}[htbp]
    \centering
    \vspace{-2mm}
    \includegraphics[width=0.9\linewidth]{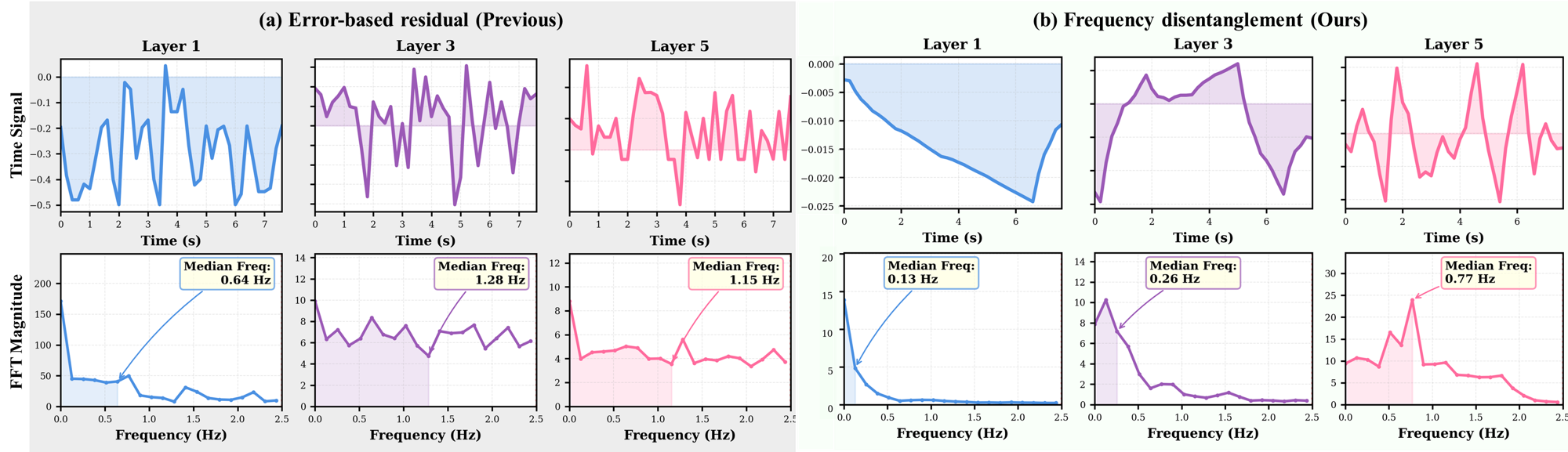}
    \caption{Comparison of time-frequency distribution. (a) RQ-VAE suffers from frequency entanglement, encoding aliased and noisy signals. (b) Our approach successfully disentangles motion into structured, hierarchical frequency bands.}
    \vspace{-3mm}
    \label{fig:intro}
\end{figure*}

\begin{figure}[t]
    \centering
    \includegraphics[width=0.9\linewidth]{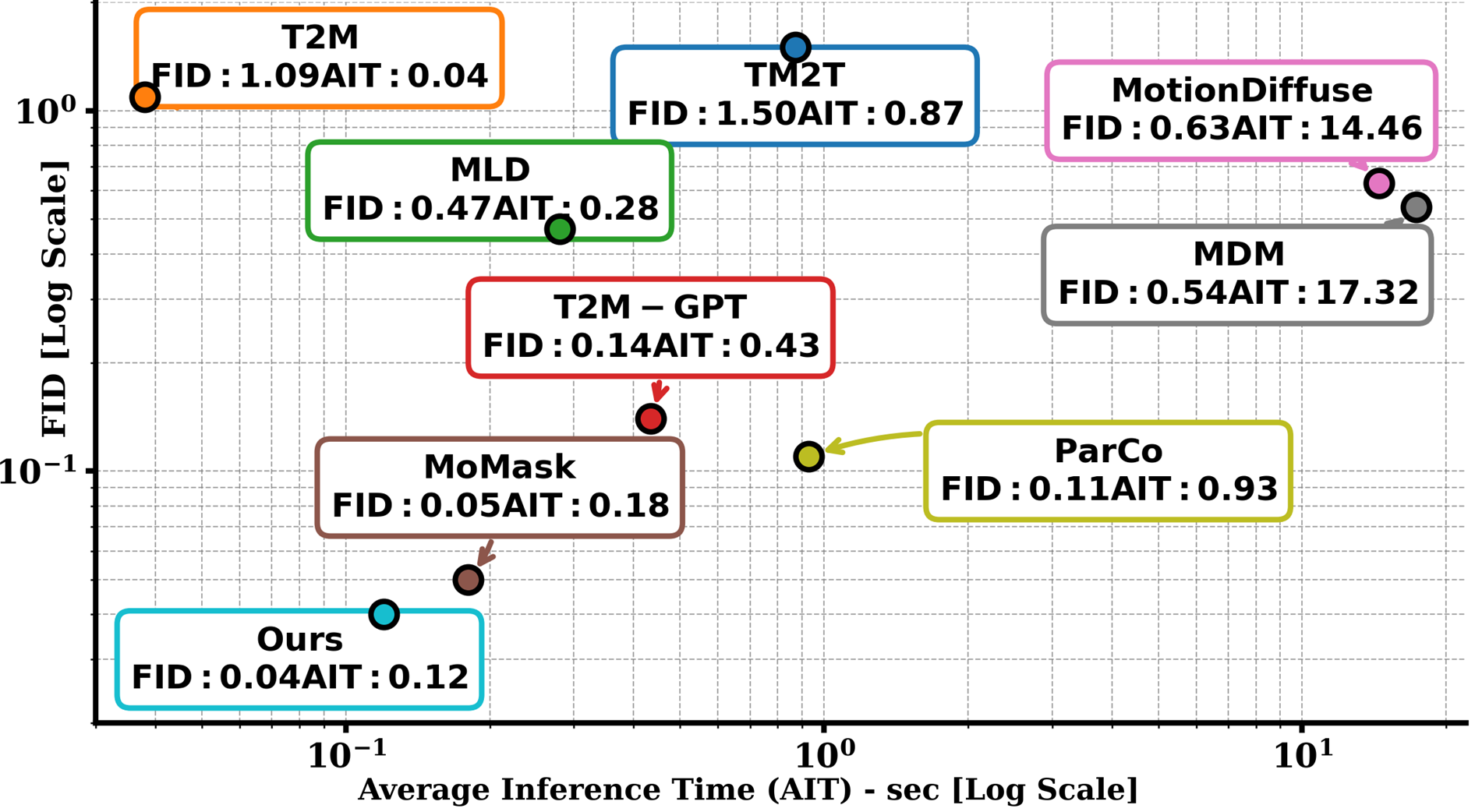}
    \vspace{-2mm}
    \caption{Quality-efficiency trade-off.}
    \vspace{-5mm}
    \label{fig:fid_time}
\end{figure}

\noindent\textbf{Spectral analysis.} Figure~\ref{fig:psd} provides direct spectral evidence. The power spectral density of motions from uniform representations rolls off sharply above 2\,Hz, indicating systematic high-frequency loss, whereas FreqMo tracks the ground-truth spectrum up to 6\,Hz. This is consistent with the lower Acceleration and Jerk Errors in Table~\ref{tab:repr_quality}.

\noindent\textbf{High-frequency preservation.} Figure~\ref{fig:acce} compares per-frame acceleration between generated and ground-truth motion. FreqMo (red) closely tracks the ground-truth peaks in timing and magnitude, capturing the abrupt transients of stumbling and diving, whereas uniform-grid baselines stay nearly flat or drift out of phase. This confirms that the frequency-disentangled representation preserves the rapid dynamics that uniform encodings attenuate.

\begin{table}[t]
\centering
\definecolor{ForestGreen}{rgb}{0.13, 0.55, 0.13} 
\definecolor{DarkRed}{rgb}{0.6, 0.0, 0.0}

\newcommand{\fidgain}[1]{{\color{ForestGreen}\footnotesize #1 $\downarrow$}}
\newcommand{\fidloss}[1]{{\color{DarkRed}\footnotesize #1 $\uparrow$}}

\resizebox{0.95\columnwidth}{!}{%
\begin{tabular}{lcccc} 
\toprule
\textbf{Level } & \textbf{ Freq. Range } & \textbf{ \#Tokens } & \textbf{ Cumulation FID } & \textbf{ Ablation FID } \\ 
\midrule
L0 (Base)  & 0.0--0.1 Hz & 4  & 6.44 \ --             & 1.30 \ \fidloss{1.26} \\
L1         & 0.0--0.2 Hz & 5  & 3.54 \ \fidgain{2.90} & 0.16 \ \fidloss{0.12} \\
L2         & 0.0--0.4 Hz & 8  & 1.49 \ \fidgain{2.05} & 0.15 \ \fidloss{0.11} \\
L3         & 0.1--0.8 Hz & 14 & 0.47 \ \fidgain{1.02} & 0.16 \ \fidloss{0.12} \\
L4         & 0.1--1.5 Hz & 26 & 0.11 \ \fidgain{0.36} & 0.19 \ \fidloss{0.15} \\
L5 (Fine)  & 0.1--2.5 Hz & 49 & 0.04 \ \fidgain{0.07} & 0.12 \ \fidloss{0.08} \\ 
\midrule
\textbf{Full (All)} & 0.0--2.5Hz & 106 & \textbf{0.04} \ -- & \textbf{0.04} \ -- \\ 
\bottomrule
\end{tabular}%
}
\vspace{-2mm}
\caption{Hierarchical frequency contribution analysis.}
\label{tab:freq_contribution}
\end{table}

\noindent\textbf{Frequency band contribution.} Table~\ref{tab:freq_contribution} quantifies the role of each band. Cumulatively adding bands from coarse to fine yields monotonic FID improvement, validating the coarse-to-fine strategy. Removing the base layer causes catastrophic degradation, confirming it captures the essential global structure. Removing any finer band causes degradation, indicating each contributes complementary high-frequency detail.

\subsection{Analysis}
\label{sec:analysis}

\begin{table}[t]
\small
\centering

\resizebox{0.9\columnwidth}{!}{%
\begin{tabular}{l@{\hspace{0.8em}}ccc|c}
\toprule
\textbf{Configuration} & \textbf{Multi-scale} & \textbf{Wavelet} & \textbf{Res.} & \textbf{FID$\downarrow$} \\
\midrule
\rowcolor{gray!15}
\multicolumn{5}{l}{\textit{Group.~1: Tokenizer Design}} \\
(a) Single-scale & \xmark & \xmark & \xmark & 0.141 \\
(b) Err. expansion & \xmark & \xmark & \cmark & 0.082 \\
(c) Time domain & \cmark & \xmark & \cmark & 0.064 \\
(d) Freq. domain & \cmark & \cmark & \xmark & 0.092 \\
\midrule
\textbf{Full (FreqMo)} & \cmark & \cmark & \cmark & \textbf{0.040} \\
\midrule
\rowcolor{gray!15}
\multicolumn{5}{l}{\textit{Group.~2: Generator Components}} \\
w/o ScAttn & \multicolumn{3}{c|}{\multirow{2}{*}{Full Model}} & 0.053 \\
Autoregressive & \multicolumn{3}{c|}{} & 0.088 \\
\midrule
\rowcolor{gray!15}
\multicolumn{5}{l}{\textit{Group.~3: Decomposition Depth $L$}} \\
$L = 1$ & \multicolumn{3}{c|}{\multirow{6}{*}{Full config.}} & 0.087 \\
$L = 2$ & \multicolumn{3}{c|}{} & 0.075 \\
$L = 3$ & \multicolumn{3}{c|}{} & 0.052 \\
$L = 4$ & \multicolumn{3}{c|}{} & 0.059 \\
$L = 5$ (default) & \multicolumn{3}{c|}{} & \textbf{0.040} \\
$L = 6$ & \multicolumn{3}{c|}{} & 0.043 \\
\midrule
\rowcolor{gray!15}
\multicolumn{5}{l}{\textit{Group.~4: Wavelet Basis}} \\
Haar (db1) & \multicolumn{3}{c|}{\multirow{4}{*}{Full config.}} & 0.059 \\
Daubechies-2 (default) & \multicolumn{3}{c|}{} & \textbf{0.040} \\
Daubechies-4 & \multicolumn{3}{c|}{} & 0.065 \\
Symlet-3 & \multicolumn{3}{c|}{} & 0.042 \\
\bottomrule
\end{tabular}%
}
\vspace{-2mm}
\caption{Ablation study. Tokenizer components (Group.~1), generator architecture (Group.~2), decomposition depth (Group.~3), and wavelet basis (Group.~4).}
\label{tab:ablation_final}

\end{table}

\textbf{Ablation study.} Table~\ref{tab:ablation_final} validates each design choice. Group~1 isolates tokenizer components: removing either wavelet decomposition (c) or residual quantization (d) substantially degrades FID, while their combination yields the best result, confirming that frequency disentanglement and progressive refinement are synergistic. Group~2 evaluates the generator: removing Scale-Aware Attention increases FID from 0.040 to 0.053, and switching to autoregressive decoding degrades it to 0.088, confirming both are essential. Group~3 shows that performance improves with decomposition depth up to $L{=}5$, beyond which it marginally degrades. Group~4 examines wavelet basis sensitivity: performance stays stable across bases, indicating the gains stem from the multi-scale paradigm rather than any specific basis.

\noindent\textbf{Frequency structure: FreqMo vs.\ RQ-VAE.} As the dominant multi-stage quantization baseline and the closest counterpart to UFRQ in token budget, RQ-VAE offers the natural comparison for internal latent organization. Figure~\ref{fig:intro} shows RQ-VAE exhibits entangled residuals with aliased spectra, each stage encoding an unstructured frequency mixture rather than isolating specific components. FreqMo instead produces cleanly separated bands, each capturing dynamics at a distinct temporal scale. This indicates UFRQ achieves genuine frequency disentanglement, not generic error reduction.

\noindent\textbf{Quality-efficiency trade-off.} Figure~\ref{fig:fid_time} plots generation quality (FID) against inference speed (AIT, average inference time) on a log scale. FreqMo sits in the optimal bottom-left quadrant, being both faster and more accurate.

\section{Conclusion}

This work introduces FreqMo, which addresses an overlooked limitation in motion generation: frequency entanglement from uniform temporal tokenization. Decomposing motion into hierarchical frequency bands and encoding them in a unified codebook via coarse-to-fine residual quantization, FreqMo achieves SOTA fidelity with faithful high-frequency preservation. Its consistent gains across discrete and continuous paradigms confirm the improvement stems from the multi-scale representation, pointing to frequency-disentangled modeling as a direction beyond human motion.

\bibliography{aaai2027}

@inproceedings{guo2022generating,
  title={Generating diverse and natural 3d human motions from text},
  author={Guo, Chuan and Zou, Shihao and Zuo, Xinxin and Wang, Sen and Ji, Wei and Li, Xingyu and Cheng, Li},
  booktitle={Proceedings of the IEEE/CVF conference on computer vision and pattern recognition},
  pages={5152--5161},
  year={2022}
}

@inproceedings{bian2025motioncraft,
  title={Motioncraft: Crafting whole-body motion with plug-and-play multimodal controls},
  author={Bian, Yuxuan and Zeng, Ailing and Ju, Xuan and Liu, Xian and Zhang, Zhaoyang and Liu, Wei and Xu, Qiang},
  booktitle={Proceedings of the AAAI Conference on Artificial Intelligence},
  volume={39},
  number={2},
  pages={1880--1888},
  year={2025}
}

@article{zhang2026motionvla,
  title={MotionVLA: Vision-Language-Action Model for Humanoid Motion},
  author={Zhang, Nonghai and Zhai, Siyu and Li, Yanjun and Zhang, Zeyu and Yin, Zhihan and Guo, Yandong and Shi, Boxin and Tang, Hao},
  journal={arXiv preprint arXiv:2606.15142},
  year={2026}
}

@inproceedings{cho2025discord,
  title={DisCoRD: Discrete Tokens to Continuous Motion via Rectified Flow Decoding},
  author={Cho, Jungbin and Kim, Junwan and Kim, Jisoo and Kim, Minseo and Kang, Mingu and Hong, Sungeun and Oh, Tae-Hyun and Yu, Youngjae},
  booktitle={Proceedings of the IEEE/CVF International Conference on Computer Vision},
  pages={14602--14612},
  year={2025}
}

@inproceedings{dai2024motionlcm,
  title={Motionlcm: Real-time controllable motion generation via latent consistency model},
  author={Dai, Wenxun and Chen, Ling-Hao and Wang, Jingbo and Liu, Jinpeng and Dai, Bo and Tang, Yansong},
  booktitle={European Conference on Computer Vision},
  pages={390--408},
  year={2024},
  organization={Springer}
}

@article{plappert2016kit,
  title={The kit motion-language dataset},
  author={Plappert, Matthias and Mandery, Christian and Asfour, Tamim},
  journal={Big data},
  volume={4},
  number={4},
  pages={236--252},
  year={2016},
  publisher={Mary Ann Liebert, Inc. 140 Huguenot Street, 3rd Floor New Rochelle, NY 10801 USA}
}

@inproceedings{yu2026cmdm,
  title={Causal Motion Diffusion Models for Autoregressive Motion Generation},
  author={Yu, Qing and Watanabe, Atsushi and Fujiwara, Kent},
  booktitle={Proceedings of the IEEE/CVF Conference on Computer Vision and Pattern Recognition (CVPR)},
  year={2026}
}

@inproceedings{xu2026omnimotionx,
  title={OmniMotion-X: Versatile Multimodal Whole-Body Motion Generation},
  author={Xu, Guowei and Bian, Yuxuan and Zeng, Ailing and Chen, Zhuo and Shi, Mingyi and Huang, Shaoli and Li, Wen and Duan, Lixin and Xu, Qiang},
  booktitle={Proceedings of the IEEE/CVF Conference on Computer Vision and Pattern Recognition (CVPR)},
  year={2026}
}

@inproceedings{weng2026realign,
  title={ReAlign: Text-to-Motion Generation via Step-Aware Reward-Guided Alignment},
  author={Weng, Wanjiang and Tan, Xiaofeng and Wang, Junbo and Xie, Guo-Sen and Zhou, Pan and Wang, Hongsong},
  booktitle={Proceedings of the AAAI Conference on Artificial Intelligence (AAAI)},
  year={2026}
}

@inproceedings{meng2025mardm,
  title={Rethinking Diffusion for Text-Driven Human Motion Generation: Redundant Representations, Evaluation, and Masked Autoregression},
  author={Meng, Zichong and Xie, Yiming and Peng, Xiaogang and Han, Zeyu and Jiang, Huaizu},
  booktitle={Proceedings of the IEEE/CVF Conference on Computer Vision and Pattern Recognition (CVPR)},
  pages={27859--27871},
  year={2025}
}

@article{fu2025mogo,
  title={MOGO: Residual Quantized Hierarchical Causal Transformer for High-Quality and Real-Time 3D Human Motion Generation},
  author={Fu, Dongjie and Sun, Tengjiao and Fang, Pengcheng and Cai, Xiaohao and Kim, Hansung},
  journal={arXiv preprint arXiv:2506.05952},
  year={2025}
}

@inproceedings{zeng2025lightt2m,
  title={Light-T2M: A Lightweight and Fast Model for Text-to-motion Generation},
  author={Zeng, Ling-An and Huang, Guohong and Wu, Gaojie and Zheng, Wei-Shi},
  booktitle={Proceedings of the AAAI Conference on Artificial Intelligence (AAAI)},
  year={2025}
}

@inproceedings{guo2024momask,
  title={Momask: Generative masked modeling of 3d human motions},
  author={Guo, Chuan and Mu, Yuxuan and Javed, Muhammad Gohar and Wang, Sen and Cheng, Li},
  booktitle={Proceedings of the IEEE/CVF Conference on Computer Vision and Pattern Recognition},
  pages={1900--1910},
  year={2024}
}

@inproceedings{lu2025scamo,
  title={ScaMo: Exploring the Scaling Law in Autoregressive Motion Generation Model},
  author={Lu, Shunlin and Wang, Jingbo and Lu, Zeyu and Chen, Ling-Hao and Dai, Wenxun and Dong, Junting and Dou, Zhiyang and Dai, Bo and Zhang, Ruimao},
  booktitle={Proceedings of the IEEE/CVF Conference on Computer Vision and Pattern Recognition (CVPR)},
  pages={27872--27882},
  year={2025}
}

@inproceedings{guo2022tm2t,
  title={Tm2t: Stochastic and tokenized modeling for the reciprocal generation of 3d human motions and texts},
  author={Guo, Chuan and Zuo, Xinxin and Wang, Sen and Cheng, Li},
  booktitle={European Conference on Computer Vision},
  pages={580--597},
  year={2022},
  organization={Springer}
}

@inproceedings{xiao2025motionstreamer,
  title={MotionStreamer: Streaming Motion Generation via Diffusion-based Autoregressive Model in Causal Latent Space},
  author={Xiao, Lixing and Lu, Shunlin and Pi, Huaijin and Fan, Ke and Pan, Liang and Zhou, Yueer and Feng, Ziyong and Zhou, Xiaowei and Peng, Sida and Wang, Jingbo},
  booktitle={Proceedings of the IEEE/CVF International Conference on Computer Vision (ICCV)},
  pages={10086--10096},
  year={2025}
}

@article{tevet2022human,
  title={Human motion diffusion model},
  author={Tevet, Guy and Raab, Sigal and Gordon, Brian and Shafir, Yonatan and Cohen-Or, Daniel and Bermano, Amit H},
  journal={arXiv preprint arXiv:2209.14916},
  year={2022}
}

@inproceedings{chen2023executing,
  title={Executing your commands via motion diffusion in latent space},
  author={Chen, Xin and Jiang, Biao and Liu, Wen and Huang, Zilong and Fu, Bin and Chen, Tao and Yu, Gang},
  booktitle={Proceedings of the IEEE/CVF conference on computer vision and pattern recognition},
  pages={18000--18010},
  year={2023}
}

@article{zhang2024motiondiffuse,
  title={Motiondiffuse: Text-driven human motion generation with diffusion model},
  author={Zhang, Mingyuan and Cai, Zhongang and Pan, Liang and Hong, Fangzhou and Guo, Xinying and Yang, Lei and Liu, Ziwei},
  journal={IEEE transactions on pattern analysis and machine intelligence},
  volume={46},
  number={6},
  pages={4115--4128},
  year={2024},
  publisher={IEEE}
}

@inproceedings{zhang2023remodiffuse,
  title={Remodiffuse: Retrieval-augmented motion diffusion model},
  author={Zhang, Mingyuan and Guo, Xinying and Pan, Liang and Cai, Zhongang and Hong, Fangzhou and Li, Huirong and Yang, Lei and Liu, Ziwei},
  booktitle={Proceedings of the IEEE/CVF International Conference on Computer Vision},
  pages={364--373},
  year={2023}
}

@inproceedings{zhang2023generating,
  title={Generating human motion from textual descriptions with discrete representations},
  author={Zhang, Jianrong and Zhang, Yangsong and Cun, Xiaodong and Zhang, Yong and Zhao, Hongwei and Lu, Hongtao and Shen, Xi and Shan, Ying},
  booktitle={Proceedings of the IEEE/CVF conference on computer vision and pattern recognition},
  pages={14730--14740},
  year={2023}
}

@book{mallat1999wavelet,
  title={A wavelet tour of signal processing},
  author={Mallat, St{\'e}phane},
  year={1999},
  publisher={Academic press}
}

@article{kingma2013auto,
  title={Auto-encoding variational bayes},
  author={Kingma, Diederik P and Welling, Max},
  journal={arXiv preprint arXiv:1312.6114},
  year={2013}
}

@inproceedings{oord2017neural,
  title={Neural discrete representation learning},
  author={Van Den Oord, Aaron and Vinyals, Oriol and Kavukcuoglu, Koray},
  booktitle={Proceedings of the 31st International Conference on Neural Information Processing Systems (NIPS)},
  pages={6309--6318},
  year={2017}
}

@inproceedings{ho2020denoising,
  title={Denoising diffusion probabilistic models},
  author={Ho, Jonathan and Jain, Ajay and Abbeel, Pieter},
  booktitle={Advances in Neural Information Processing Systems (NeurIPS)},
  volume={33},
  pages={6840--6851},
  year={2020}
}

@article{jiang2023motiongpt,
  title={Motiongpt: Human motion as a foreign language},
  author={Jiang, Biao and Chen, Xin and Liu, Wen and Yu, Jingyi and Yu, Gang and Chen, Tao},
  journal={Advances in Neural Information Processing Systems},
  volume={36},
  pages={20067--20079},
  year={2023}
}

@inproceedings{chang2022maskgit,
  title={Maskgit: Masked generative image transformer},
  author={Chang, Huiwen and Zhang, Han and Jiang, Lu and Liu, Ce and Freeman, William T},
  booktitle={Proceedings of the IEEE/CVF conference on computer vision and pattern recognition},
  pages={11315--11325},
  year={2022}
}

@inproceedings{devlin2019bert,
  title={Bert: Pre-training of deep bidirectional transformers for language understanding},
  author={Devlin, Jacob and Chang, Ming-Wei and Lee, Kenton and Toutanova, Kristina},
  booktitle={Proceedings of the 2019 conference of the North American chapter of the association for computational linguistics: human language technologies, volume 1 (long and short papers)},
  pages={4171--4186},
  year={2019}
}

@inproceedings{vaswani2017attention,
  title={Attention is all you need},
  author={Vaswani, Ashish and Shazeer, Noam and Parmar, Niki and Uszkoreit, Jakob and Jones, Llion and Gomez, Aidan N and Kaiser, {\L}ukasz and Polosukhin, Illia},
  booktitle={Advances in Neural Information Processing Systems (NIPS)},
  pages={5998--6008},
  year={2017}
}

@inproceedings{zou2024parco,
  title={Parco: Part-coordinating text-to-motion synthesis},
  author={Zou, Qiran and Yuan, Shangyuan and Du, Shian and Wang, Yu and Liu, Chang and Xu, Yi and Chen, Jie and Ji, Xiangyang},
  booktitle={European Conference on Computer Vision},
  pages={126--143},
  year={2024},
  organization={Springer}
}

@inproceedings{radford2021learning,
  title={Learning transferable visual models from natural language supervision},
  author={Radford, Alec and Kim, Jong Wook and Hallacy, Chris and Ramesh, Aditya and Goh, Gabriel and Agarwal, Sandhini and Sastry, Girish and Askell, Amanda and Mishkin, Pamela and Clark, Jack and others},
  booktitle={International Conference on Machine Learning (ICML)},
  pages={8748--8763},
  year={2021},
  organization={PMLR}
}

@article{yuan2024mogents,
  title={Mogents: Motion generation based on spatial-temporal joint modeling},
  author={Yuan, Weihao and He, Yisheng and Shen, Weichao and Dong, Yuan and Gu, Xiaodong and Dong, Zilong and Bo, Liefeng and Huang, Qixing},
  journal={Advances in Neural Information Processing Systems},
  volume={37},
  pages={130739--130763},
  year={2024}
}

@article{fu2024mogo,
  title={Mogo: RQ Hierarchical Causal Transformer for High-Quality 3D Human Motion Generation},
  author={Fu, Dongjie},
  journal={arXiv preprint arXiv:2412.07797},
  year={2024}
}

@article{allen1977unified,
  title={A unified approach to short-time Fourier analysis and synthesis},
  author={Allen, Jont B and Rabiner, Lawrence R},
  journal={Proceedings of the IEEE},
  volume={65},
  number={11},
  pages={1558--1564},
  year={1977},
  publisher={IEEE}
}

@book{fourier1822theorie,
  title={Th{\'e}orie analytique de la chaleur},
  author={Fourier, Jean-Baptiste-Joseph},
  year={1822},
  publisher={Paris: Firmin Didot, p{\`e}re et fils}
}

@article{khani2025motionsurvey,
  title={Motion Generation: A Survey of Generative Approaches and Benchmarks},
  author={Khani, Aliasghar and Rampini, Arianna and Roy, Bruno and Nadela, Larasika and Kaplan, Noa and Atherton, Evan and Cheung, Derek and Bibliowicz, Jacky},
  journal={arXiv preprint arXiv:2507.05419},
  year={2025}
}

@book{burdea2003virtual,
  title={Virtual reality technology},
  author={Burdea, Grigore C and Coiffet, Philippe},
  year={2003},
  publisher={John Wiley \& Sons}
}

@article{duan2022survey,
  title={A survey of embodied ai: From simulators to research tasks},
  author={Duan, Jiafei and Yu, Samson and Tan, Hui Li and Zhu, Hongyuan and Tan, Cheston},
  journal={IEEE Transactions on Emerging Topics in Computational Intelligence},
  volume={6},
  number={2},
  pages={230--244},
  year={2022},
  publisher={IEEE}
}

@inproceedings{varol2017learning,
  title={Learning from synthetic humans},
  author={Varol, Gul and Romero, Javier and Martin, Xavier and Mahmood, Naureen and Black, Michael J and Laptev, Ivan and Schmid, Cordelia},
  booktitle={Proceedings of the IEEE conference on computer vision and pattern recognition},
  pages={109--117},
  year={2017}
}

@article{liu2022investigating,
  title={Investigating pose representations and motion contexts modeling for 3D motion prediction},
  author={Liu, Zhenguang and Wu, Shuang and Jin, Shuyuan and Ji, Shouling and Liu, Qi and Lu, Shijian and Cheng, Li},
  journal={IEEE transactions on pattern analysis and machine intelligence},
  volume={45},
  number={1},
  pages={681--697},
  year={2022},
  publisher={IEEE}
}

@inproceedings{siarohin2021motion,
  title={Motion representations for articulated animation},
  author={Siarohin, Aliaksandr and Woodford, Oliver J and Ren, Jian and Chai, Menglei and Tulyakov, Sergey},
  booktitle={Proceedings of the IEEE/CVF conference on computer vision and pattern recognition},
  pages={13653--13662},
  year={2021}
}

@inproceedings{zhu2023motionbert,
  title={Motionbert: A unified perspective on learning human motion representations},
  author={Zhu, Wentao and Ma, Xiaoxuan and Liu, Zhaoyang and Liu, Libin and Wu, Wayne and Wang, Yizhou},
  booktitle={Proceedings of the IEEE/CVF international conference on computer vision},
  pages={15085--15099},
  year={2023}
}

@inproceedings{pinyoanuntapong2024bamm,
  title={Bamm: Bidirectional autoregressive motion model},
  author={Pinyoanuntapong, Ekkasit and Saleem, Muhammad Usama and Wang, Pu and Lee, Minwoo and Das, Srijan and Chen, Chen},
  booktitle={European Conference on Computer Vision},
  pages={172--190},
  year={2024},
  organization={Springer}
}

@inproceedings{peebles2023scalable,
  title={Scalable diffusion models with transformers},
  author={Peebles, William and Xie, Saining},
  booktitle={Proceedings of the IEEE/CVF international conference on computer vision},
  pages={4195--4205},
  year={2023}
}

@article{achiam2023gpt,
  title={Gpt-4 technical report},
  author={Achiam, Josh and Adler, Steven and Agarwal, Sandhini and Ahmad, Lama and Akkaya, Ilge and Aleman, Florencia Leoni and Almeida, Diogo and Altenschmidt, Janko and Altman, Sam and Anadkat, Shyamal and others},
  journal={arXiv preprint arXiv:2303.08774},
  year={2023}
}

\end{document}